\documentclass[acmsmall]{acmart}
\usepackage{booktabs} 
\usepackage{subfigure}
\usepackage{blindtext}
\usepackage{graphicx}
\usepackage{csquotes}
\usepackage{caption}
\usepackage{amsmath}
\usepackage{siunitx}
\usepackage{microtype}
\usepackage{multicol}
\usepackage{upgreek}
\usepackage{booktabs,tabularx,ragged2e,url}
\usepackage[english]{babel}
\usepackage{fancyhdr}
\usepackage[noend]{algpseudocode}
\usepackage{float}
\usepackage{blindtext}

\usepackage{academicons}
\usepackage{xcolor}

\usepackage{nomencl}
\makenomenclature

\usepackage[ruled,linesnumbered,commentsnumbered]{algorithm2e}

\SetAlFnt{\small}
\SetAlCapFnt{\small}
\SetAlCapNameFnt{\small}
\SetAlCapHSkip{0pt}
\IncMargin{-\parindent}

\usepackage[hang,flushmargin]{footmisc}

\usepackage{xcolor,colortbl}
\definecolor{Gray}{gray}{0.90}
\newcolumntype{a}{>{\columncolor{Gray}}c}

\usepackage{xcolor,soul}
\definecolor{light-gray}{gray}{0.95}
\sethlcolor{light-gray}

\graphicspath{{Figures/}}
\newcommand{\fig}[1]{Fig.~\ref{#1}}
\newcommand{\tb}[1]{Table~\ref{#1}}
\newcommand{\eq}[1]{(\ref{#1})}

\usepackage{enumitem,booktabs}
\usepackage[referable]{threeparttablex}
\renewlist{tablenotes}{enumerate}{1}
\makeatletter
\setlist[tablenotes]{label=\tnote{\alph*},ref=\alph*,itemsep=\z@,topsep=\z@skip,partopsep=\z@skip,parsep=\z@,itemindent=\z@,labelsep=.2em,leftmargin=*,align=left,before={\footnotesize}}
\makeatother

\usepackage[font={small}]{caption}

\DeclareCaptionLabelSeparator{periodspace}{.\quad}
\usepackage{enumitem}
\newlist{todolist}{itemize}{2}
\setlist[todolist]{label=$\square$}
\usepackage{pifont}
\usepackage[commandnameprefix=always]{changes}

\AtBeginDocument{%
  }

\setcopyright{acmlicensed}

\begin{document}

\title{Minion Gated Recurrent Unit for Continual Learning \vspace{3mm}}

\author{Abdullah M. Zyarah}
\email{abdullah.zyarah@uob.edu.iq}
\orcid{0000-0001-8220-5285}
\authornotemark[1]
\affiliation{%
  \institution{\\NuAI Lab, Department of Electrical and Computer Engineering, University of Texas at San Antonio, USA \\ Department of Electrical Engineering, University of Baghdad. }
  \country{Iraq}
}

\author{Dhireesha Kudithipudi}
\orcid{0000-0003-4462-5224}
\affiliation{%
  \institution{\\NuAI Lab, Department of Electrical and Computer Engineering, University of Texas at San Antonio}
  \country{USA}}
\email{dk@utsa.edu}

\begin{abstract}

The increasing demand for continual learning in sequential data processing has led to progressively complex training methodologies and larger recurrent network architectures. Consequently, this has widened the knowledge gap between continual learning with recurrent neural networks (RNNs) and their ability to operate on devices with limited memory and compute. To address this challenge, we investigate the effectiveness of simplifying RNN architectures, particularly gated recurrent unit (GRU), and its impact on both single-task and multitask sequential learning. We propose a new variant of GRU, namely the minion recurrent unit (MiRU). MiRU replaces conventional gating mechanisms with scaling coefficients to regulate dynamic updates of hidden states and historical context, reducing computational costs and memory requirements. Despite its simplified architecture, MiRU maintains performance comparable to the standard GRU while achieving 2.90$\times$ faster training and reducing parameter usage by 2.88$\times$, as demonstrated through evaluations on sequential image classification and natural language processing benchmarks. The impact of model simplification on its learning capacity is also investigated by performing continual learning tasks with a rehearsal-based strategy and global inhibition. We find that MiRU demonstrates stable performance in multitask learning even when using only rehearsal, unlike the standard GRU and its variants. These features position MiRU as a promising candidate for edge-device applications.


\end{abstract}




\maketitle

\section{Introduction}
\label{sec:introduction}
The recent years have witnessed continual learning as an integral aspect of modern machine learning models, particularly for systems deployed in dynamic environments, where models must continuously learn and handle changes in data distributions without forgetting previously learned experiences~\cite{kudithipudi2023design, silver2013lifelong}. Obtaining such an objective is highly challenging due to the problem of catastrophic forgetting \cite{mcclelland2020integration, grossberg1982does}. Catastrophic forgetting typically occurs in machine learning models due to overlapping internal representations~\cite{wickramasinghe2023continual}. In such cases, the model overwrites previously learned parameters with new ones, reflecting the data distribution of the current task. To address this problem, several solutions have been presented in the literature, including complex synaptic connections or regularization-based strategies~\cite{smith2023closer, liu2018rotate}, dynamic architectures~\cite{rusu2016progressive, von2019continual}, or rehearsal-based
strategies (replay mechanisms)~\cite{silver2002task, shin2017continual}. 
However, most of these solutions have been extensively investigated in the context of spatial tasks and reinforcement learning settings~\cite{dohare2024loss, wang2024comprehensive}. Temporal tasks, particularly those processed by recurrent neural networks (RNNs), are yet to be fully explored~\cite{cossu2021continual}. 

Recurrent neural networks, such as long-short-term memory (LSTM)~\cite{hochreiter1997long} and gated-recurrent unit (GRU)~\cite{cho2014learning}, have demonstrated exceptional performance in a wide range of applications, including natural language processing~\cite{wasef2023soc}, time series data analysis and forecasting~\cite{zhang2024novel}, robotics and autonomous systems~\cite{wang2024cle}. However, their complex architecture and high parameter count make them less attractive for continual learning and sequence classification on edge devices, where computational and memory resources are limited. To this end, several variants of LSTM and GRU have been presented in the literature with the aim of reducing the number of parameters and simplifying the network architecture, thereby minimizing the storage requirements and computational costs. Some of these variants place emphasis on simplifying the structure of the gates within LSTM or GRU units, which control the flow of information and local storage, by techniques such as combining signals, pruning connections, etc.~\cite{heck2017simplified, dey2017gate}. Others adopt a more direct approach by merging the gates themselves to create a more compact design~\cite{zhou2016minimal}. Although these approaches offer reductions in complexity, they focus on static tasks rather than adapting to dynamic and evolving tasks, where continual learning is critical. 

To enable learning dynamic and evolving tasks on resource-constrained devices, we propose a new variant of the GRU, termed the minion recurrent unit (MiRU). MiRU eliminates the reset and forget gates, replacing them with scaling coefficients, which are hyper-parameters used to control the flow of information and local storage. The proposed model significantly reduces computational cost and storage requirements during both training and inference. During evaluation, MiRU is initially verified in sequence classification tasks using the MNIST and IMDB datasets, resulting in performance comparable to the standard GRU while using approximately 2.88$\times$ fewer parameters. Then, the impact of model simplification on its learning capacity is investigated by evaluating MiRU in a domain-incremental learning (DIL) scenario~\cite{van2019three, cossu2021continual}, which mimics real-world dynamic environments. To mitigate catastrophic forgetting, we incorporate a rehearsal-based strategy in which previously seen data is stored and revisited while learning new tasks. The rehearsal-based strategy is then complemented by a biologically inspired mechanism called global inhibition, which regulates neuronal activities across tasks. The list of contributions of this work can be summarized as follows.
\begin{itemize}
    \item Introducing the minion recurrent unit (MiRU) that offers performance comparable to standard GRU and faster training while using $\sim$2.88$\times$ fewer parameters.
    \item Endow the proposed model with the capability to perform continual learning tasks through replay, while enhancing the learning stability and resource efficiency using biologically inspired mechanisms, such as global inhibition.
    \item Evaluating MiRU against several metrics, including classification accuracy, mean accuracy across tasks, latency, resource utilization, and energy consumption.
\end{itemize}

The remainder of the paper is organized as follows. Section II discusses the theory of GRU RNN and continual learning. Section III presents the minion recurrent unit. The evaluation benchmark and experimental setup are introduced in Section IV. Sections V and VI, respectively, discuss the results and conclude the paper. 

\begin{figure*}[h!t]
    \centering
    \includegraphics[width=1\textwidth]{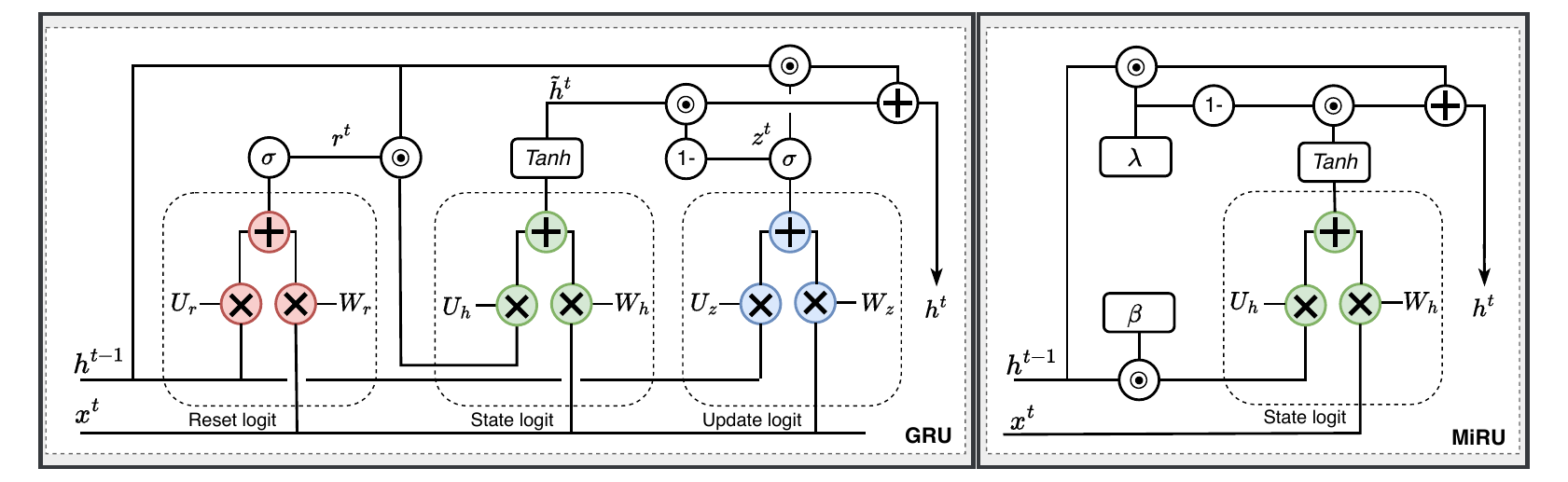}
    \caption{The data flow and operations of the standard GRU (Left) which comprises of a reset gate, an update gate, and an additional unit to compute the candidate hidden state. The proposed MiRU unit (right) which uses reset ($\beta$) and update ($\lambda$) coefficients rather than gates to control network dynamics and information updates, respectively.}
    \label{gru_arch}
\end{figure*}

\section{Background}
\subsection{Gated Recurrent Unit RNN}
The gated recurrent unit (GRU) RNN was introduced by Cho et al. in 2014~\cite{cho2014learning} as a simplified version of the LSTM. The network can adaptively capture short- and long-term dependencies~\cite{chung2014empirical} and offers performance comparable to LSTM in a diverse range of real-world applications, including learning symbolic sequences~\cite{cahuantzi2023comparison}, pandemic and cryptocurrency prices forecasting~\cite{prakash2023forecasting, seabe2023forecasting}, and emotion recognition~\cite{priyadarshini2023emotion}. 
The general architecture of the GRU RNN consists of three consecutive main layers: input, hidden, and output. These layers are densely or sparsely interconnected via weighted synaptic connections stemming from adjacent preceding layers (forward connections) or from the layer itself (recurrent connections). Based on the neuronal type forming the layer and its location, each layer in the network serves different purposes. The input layer, which is formed by linear neurons, typically serves as a buffer projecting the input features to the hidden layer through forward-weighted synaptic connections. In the hidden layer, all the salient features are extracted via one or more layers of stacked hidden neurons, namely GRU. The GRUs are stimulated by forward connections via input layer activations ($x \in \mathbb{R}^{n_x \times n_T}$)\footnote{Here, we are assuming all inputs have a fixed sequence length, $n_T$.} and lateral connections via layer activation from previous time step ($h^{t-1} \in \mathbb{R}^{n_h \times 1}$), where $n_x$ is the number of input features, $n_h$ is the number of the hidden units, and $n_T$ denotes the length of each input in time, i.e. sequence length.

Each GRU has two independent gates: reset and update, which govern the flow of information and local storage, see~\fig{gru_arch}-left. These gates are activated at different paces to capture short- and long-term dependencies. For example, units in which the reset gate is activated more frequently than the update gate tend to capture short-term dependencies and vice versa~\cite{ravanelli2018light}. However, the activation of each gate is determined by the external stimulus of the gate, see \eq{reset_g1} and ~\eq{update_g1}, where $r^t$ and $z^t$ are the activations of the reset and update gates at time $t$, $\sigma$ is the logistic activation function (sigmoid), and $W \in \mathbb{R}^{n_h \times n_x}$ and $U \in \mathbb{R}^{n_h \times n_h}$, respectively, represent the weights of forward and recurrent synaptic connections which are learned. 

\begin{equation}
    r^t = \sigma(W_r x ^ t + U_r h^{t-1} + b_r)
\label{reset_g1}
\end{equation}
\begin{equation}
    z^t = \sigma(W_z x ^ t + U_z h^{t-1} + b_z)
\label{update_g1}
\end{equation}

When inputs are presented to GRU units, the output of each gate is initially determined. Then, the outputs of the reset gates are multiplied (Hadamard multiplication) by the hidden layer activations from the previous time step ($h^{t-1}$). This step specifies the amount of information to be forgotten from the history, which may not be necessary or critical to make future decisions (i.e. irrelevant information). Thus, when $r$ is set close to zero, the hidden activation is determined entirely based on the present input, ignoring the previous hidden state. Following that, the candidate of the hidden layer activation ($\tilde{h}^t$) is calculated based on both the history and the present input as given in~\eq{h_tilde_g1}.  

\begin{equation}
    \tilde{h}^t = tanh[W_h x ^ t + U_h (r^t \odot h^{t-1}) + b_h]
\label{h_tilde_g1}
\end{equation}
\begin{equation}
    h^t = z^t \odot h^{t-1} + (1 - z^t) \odot \tilde{h}^{t}
\label{h_out}
\end{equation}
\begin{equation}
    \hat{y}^t = \sigma(W_y~ h^{t}) 
\label{yhat}
\end{equation}

After determining the hidden state candidate, the update gate controls the amount of information from the previous hidden activation that will be incorporated into the current hidden state. This is critical for retaining long-term information. However, the hidden layer activation is computed as in~\eq{h_out}, which is forwarded to the output layer with $n_y$ neurons via weighted synaptic connections between the hidden and output layers, $W_y \in \mathbb{R}^{n_h \times n_y}$, to get classified, see~\eq{yhat}.

\subsection{GRU RNN Simplification and Optimization}
{The GRU RNN is presented as a simplified version of the LSTM that offers comparable performance in numerous real-world tasks. Nevertheless, the GRU still has a complex architecture because it is equipped with two gates and has 3-folds of parameters compared to vanilla RNNs~\cite{dey2017gate, sodhani2019toward, zyarah2023reservoir}. In the literature, several successful attempts have been made in order to further simplify the GRU RNNs. These attempts can be classified into three categories: i) gate and architecture simplification, ii) low-resource optimization, and iii) feature reduction. Regarding the gate and architecture simplification, this approach focuses on merging or simplifying the general structure of the gates within the GRU unit. For instance, Zhou et al. proposed a minimal gated unit (MGU) for recurrent neural networks~\cite{zhou2016minimal}. The MGU replaces the two gates in GRU, reset and update, with one universal gate called the forget gate, thereby reducing the number of parameters without sacrificing accuracy when verified on MNIST and Reuters newswire topics (RNT) datasets. Ravanelli et al. introduced the light gated recurrent unit (Li-GRU) that eliminates the reset gate to get more compact model~\cite{ravanelli2018light}. The authors also used $ReLU$ activation with batch normalization instead of $tanh$ to learn long-term dependencies in speech recognition applications. Inspired by the LSTM unit simplification suggested in~\cite{lu2017simplified}, Dey et al. introduced a simplified version of the standard GRU that eliminates combinations of input signals, biases, or hidden unit signals from the GRU units. The proposed design was benchmarked using MNIST and IMDB datasets and achieved performance comparable to that of the standard GRU in some scenarios. Then, the same group applied similar model reduction techniques to the MGU and verified their operation using the MNIST and RNT datasets~\cite{heck2017simplified}. Later, Yigit et al. proposed another variant of GRU that incorporates scaling factors into the reset and update gates~\cite{yiugit2021simple}. These factors control the degree of change in the hidden state caused by the input and recurrent connections. This variant was intended to enhance the performance of the GRU rather than reduce its complexity. Recently, Fathi suggested a new variant for the GRU unit, called slim GRU (SGRU), to reduce the number of parameters~\cite{salem2022gated}. The author pruned the input signals from the reset and update gates and used shared recurrent connection weights to estimate their activations.

When it comes to low-resource optimization, it encompasses techniques such as quantization and analytical computation. An example is the CryptoGRU presented by Feng et al. for low latency secure inference~\cite{feng2021cryptogru}. The presented model used homographic encryption to process linear operations such as, multiplications and additions, and adopted garbled circuits to realize non-linear operations such as, ReLU, tanh, and sigmoid activation functions. Furthermore, the authors replaced the tanh activation with ReLU and quantized both ReLU and sigmoid activations to a smaller bitwidth to accelerate the computations in the GRU unit. The proposed model was verified with IMDB and Yelp Review. Regarding the last approach, feature reduction, it involves methods like compressing features. For instance, the study by Zulqarnain et al. proposed reducing the number of parameters in GRU through the use of an auto-encoder. The auto-encoder reduces the dimensionality of the input features when the proposed model was verified with word embedding for sentiment classification~\cite{zulqarnain2020improved}. Cao et al., used feature compression and frequency band division to reduce the network parameters and computational load in GRU. The proposed model was evaluated and validated using TIMIT dataset~\cite{cao2024beamforming}. 

To ensure a proper and fair comparison between the proposed model, MiRU, and the GRU along with its variants presented in the literature, it is essential to focus on models that directly address the architectural complexity of the GRU. Specifically, the models that fall under the first category, as they aim to simplify the unit structure and align with the proposed model's objective of reducing parameters and computational overhead without compromising performance when evaluated using standard benchmarks.} 

\subsection{Continual Learning in RNNs}
In continual learning scenarios, RNNs are prone to forgetting previously learned information upon acquiring new knowledge, a problem known as catastrophic forgetting, or catastrophic inference~\cite{cossu2021continual}. It typically occurs due to the lack of balance between the stability-plasticity aspects, where plasticity indicates the network's capability to learn novel information, and stability reflects its ability to retain previously learned knowledge~\cite{riemer2018learning}. Several studies in the literature have tackled the catastrophic forgetting problem by applying common techniques, such as regularization, dynamic architecture, and replay~\cite{kudithipudi2023design}. Despite the fact that most of these techniques are not designed to handle sequence data, some are still feasible to be used with RNNs~\cite{sodhani2019toward}. For instance, regularization may be used to safeguard memories and improve stability via penalizing critical synaptic connections that are deemed important to previously learned tasks. Elastic weight consolidation (EWC)~\cite{kirkpatrick2017overcoming}, synaptic intelligence (SI)~\cite{zenke2017continual}, and learning without forgetting (LwF)~\cite{li2017learning} are commonly used representatives of the regularization approach. The dynamic architecture can be leveraged to modify the network structure by adding or removing neurons and synaptic pathways to accommodate new knowledge and regulate stability-plasticity balance~\cite{wickramasinghe2023continual}. In case of replay, it is a rehearsal-based strategy that involves fine-tuning network parameters via persisting selected samples from previously learned tasks stored in a buffer. Unlike regularization and dynamic architecture, replay is considered the most effective technique in continual learning~\cite{sodhani2019toward}. Therefore, in this work, we will mainly focus on the replay, particularly reservoir replay, and occasionally use it with biologically inspired mechanisms, such as global inhibition~\cite{zyarah2020neuromorphic}, to reduce the number of samples stored in the replay buffer and mitigate catastrophic forgetting. The global inhibition involves representing each input by a subset of active GRU units based on the neuronal activation level ($h^t$) and the desired level of sparsity ($\eta$). Given $n_h$ hidden neurons, the subset of GRU neurons to represent the input is given by~\eq{inhibition}, where $kmax$ is a function that implements the \textit{k-winner-take-all}. 
\begin{equation}
    \hat{h}^t = kmax(h^t, \eta, n_h)
\label{inhibition}
\end{equation}


\section{Minion Recurrent Unit (MiRU)}
There are several variants of GRU RNN that have been presented in the literature with the aim of minimizing the storage requirements and computational costs, thereby enabling on edge streaming data processing. In this work, we introduce new variants of GRU which mainly focuses on reducing the number of parameters, computational cost, and simplifying the learning process. The proposed variant, namely minion recurrent unit (MiRU), comes in two versions. The first version, namely MiRU-1, suggests keeping the reset gate and replacing the update gate with an update coefficient, $\lambda$. The update coefficient controls the dynamic update of the hidden states, producing a dynamical system on multiple time scales\footnote{The multiple time scales condition holds true when the update coefficients are set randomly i.e. each MiRU has its own update coefficient.}. The output of MiRU-1 can be determined using the following set of equations.

\begin{equation}
    r^t = \sigma(W_r x ^ t + U_r h^{t-1} + b_r)
\label{resetx}
\end{equation}
\begin{equation}
    \tilde{h}^t = tanh(W_h x ^ t + U_h (r^t \odot h^{t-1}) + b_h)
\label{h_tilde}
\end{equation}
\begin{equation}
    h^t = \lambda \odot h^{t-1} + (1 - \lambda) \odot \tilde{h}^{t}
\label{h_out_MiRU}
\end{equation}

The second version, called MiRU-2, goes further by also eliminating the reset gate and replacing it with a reset coefficient, $\beta$. The reset coefficient determines how much of the previous hidden state should be forgotten or reset before combining it with the new input. The hidden unit activation of MiRU-2 can be computed using the following set of equations:

\begin{equation}
    \tilde{h}^t = tanh(W_h x ^ t + U_h (\beta \odot h^{t-1}) + b_h)
\label{h_tilde_MiRU_2}
\end{equation}
\begin{equation}
    h^t = \lambda \odot h^{t-1} + (1 - \lambda) \odot \tilde{h}^{t}
\label{h_out_MiRU_2}
\end{equation}

{Both the reset and update coefficients are predefined hyperparameters that are initialized to configure the learning process in MiRU. Smaller values of update coefficient, $\lambda$, prioritize incorporating new information, while larger values emphasize retaining past information. Similarly, a higher reset coefficient, $\beta$, value encourages retention of prior hidden states, while a lower value promotes resetting and adaptation to new inputs. By carefully selecting the values of the reset and update coefficients, these scaling factors can function comparably to the update and reset gates in the standard GRU. They strike a balance between preserving past information and incorporating new data, enabling MiRU to effectively capture both short- and long-term dependencies. }

One may observe from the MiRU equations that there is a significant reduction in the number of components and parameters forming each unit as compared to the standard GRU, see~\fig{gru_arch}-right. Besides the fact that such simplification facilitates the future understanding of the fundamental learning mechanisms in RNNs, it also means a significant reduction in the storage requirements and computational complexity during training and inference. This will eventually enable porting and training RNNs locally on edge devices rather than the cloud, solving problems related to network adaptation, response time, and concerns associated with safety and privacy.

\begin{table}[!hb]
\footnotesize
\caption{The hyperparameters used in the standard GRU, its variants, and the proposed MiRU for sequence classification and domain-incremental continual learning.}
\label{model_parameters}
\begin{center}
\begin{threeparttable}
\begin{tabular}{lccc}
\toprule                     
\textbf{Parameter}  &  \textbf{MNIST} &   \textbf{IMDB} & \textbf{DIL-MNIST} \\ 
\midrule
Model size  & 28$\times$128$\times$10 & 128$\times$128$\times$1 & 28$\times$256$\times$10  \\
Optimizer & RMSPprop &  RMSPprop & Adam  \\
Learning rate & 0.001 &  0.001 & 0.001 \\
Regular.  param. & -  & 0.0006 & - \\
Mini-batch ($n_b$) &  32 & 64 & 32  \\
Epochs & 15 & 15 &  10\\
Reset coeff. & 0.55 & 0.55  & -\tnotex{tnote:robots-a1} \\
Update coeff. & 0.8 & 0.7  & -\tnotex{tnote:robots-a1} \\
Replay buffer & - & -  &  1875, 3750\tnotex{tnote:robots-a2} \\
Sparsity ($\eta$) & - & -  & 75\%\\
\midrule
\bottomrule
  \end{tabular}
      \begin{tablenotes}
      \item\label{tnote:robots-a1} Set to be random during the DIL task to capture features across mutliple time scales. 
      \item\label{tnote:robots-a2} The replay buffer size per task is initially set to 1875 then 3750 samples, respectively.
    \end{tablenotes}
\end{threeparttable}
\end{center}
\end{table}

\section{Experimental Setup}
\subsection{Benchmarks}
In order to assess the performance of the proposed MiRU model for time series classification, IMDB and MNIST datasets are used. The IMDB is a movie review dataset for binary sentiment classification~\cite{maas2011learning}. The dataset contains 50,000 highly polar movie reviews split into training and test sets. Each review is represented with 128 of the most common words and presented to RNN models as a sequence of numerical vectors of size 128 (input feature size). The MNIST dataset contains 60,000 training examples and 10,000 test examples of gray-scale hand-written digits~\cite{MNIST}. Each example, 28x28 pixels, are presented to the RNN models in form of sequences, where each row is considered as a feature vector of length 28. For the continual learning tasks, we constructed a unique sequence of $T$ tasks \{$\mathcal{D}_1, \mathcal{D}_2, .... \mathcal{D}_T$\}, derived from the dataset, known as permuted MNIST~\cite{goodfellow2013empirical}. For each task, $\mathcal{D}_t$, a fixed random permutation are generated to shuffle the pixels of images within the set while preserving the associated target label, such that the input distribution of each task is independent and all tasks will be of equal difficulty.



\subsection{Setup \& Evaluation}
As alluded to earlier, the proposed MiRU is evaluated on sequence classification tasks and domain-incremental continual learning scenario. To ensure consistent evaluation and conclusions with previous works~\cite{heck2017simplified, dey2017gate, zhou2016minimal}, we attempt to use approximately similar network architectures, hyperparameters, and optimizers (see~\tb{model_parameters}). During sequence classification of IMDB dataset, the network is set up to have 128 neurons in the hidden layer and 1 neuron in the output to identify positive and negative comments. The reset and update coefficients are set to 0.55 and 0.7, respectively, for all MiRU units. The training is done using RMSProp with learning rate of 1e-3 and a mini-batch ($n_b$) of 64. The same setup is also employed for other networks, GRU and its variants, to ensure a fair comparison. For MNIST dataset, the same number of hidden neurons is used, but 10 neurons are employed in the output layer corresponding to each class. The training is also done using RMSProp with learning rate of 1e-3 and a mini-batch of 32. During the DIL, we expand the network architecture to have 256 neurons in the hidden layer to improve its learning capacity. Furthermore, we change the network hyperparameters and switch to Adam optimizer to account for task learning complexity and accelerate the learning process. Replay and global inhibition mechanisms are also incorporated to alleviate the forgetting. The replay involves interleaving randomly selected samples from the replay buffer ${M}$ with the training of the current examples. In every task, $t \in T$, the buffer is iteratively updated with ${k}$ samples chosen from $j^{th}$ mini-batch $\mathcal{B}_{jt}$ within $\mathcal{D}_t$ as in~\eq{mem_buff} and~\eq{batch}. 
\begin{equation}
    'M = M \cup \{(x_i,y_i)\sim \mathcal{B}_{jt}\}^{k}_{i=1} 
\label{mem_buff}
\end{equation}
\begin{equation}
    \mathcal{B}_{jt} = \{(x_i,y_i)\sim \mathcal{D}_t\}^{n_b}_{i=1} 
\label{batch}
\end{equation}

During interleaving, which occurs while learning subsequent tasks, from $t+1$ to $T$, ${k}$ samples are chosen from $M$ to be augmented with each mini-batch. Given the current task, $t+1$, and the $j^{th}$ mini-batch, the samples selected from $M$ are augmented with $\mathcal{B}_{jt+1}$ as given in~\eq{batch_new} and \eq{batch_old}.
\begin{equation}
    '\mathcal{B}_{jt+1} = \mathcal{B}_{jt+1} \cup \{(x_i,y_i)\sim {M}\}^{k}_{i=1} 
\label{batch_new}
\end{equation}
\begin{equation}
    \mathcal{B}_{jt+1} = \{(x_i,y_i)\sim \mathcal{D}_{t+1}\}^{n_b}_{i=1} 
\label{batch_old}
\end{equation}

The samples that are selected from the mini-batch to update the replay buffer, and from the buffer to be augmented with the mini-batches during interleaving, are chosen using reservoir sampling~\cite{vitter1985random}. Reservoir sampling enables selecting examples from non-stationary streams of unknown length with equal probabilities and makes efficient use of the limited memory buffer~\cite{zhuo2023continual}. When it comes to the global inhibition, it is incorporated in the hidden layer and applied by creating a binary mask to be multiplied by the neuronal activations, $h^t$. The mask is generated in each time step with ones indexing the neurons with the highest activation levels and zeros elsewhere.

\begin{figure*}[!t]
\centering
\subfigure{\includegraphics[width=45mm, height=40mm]{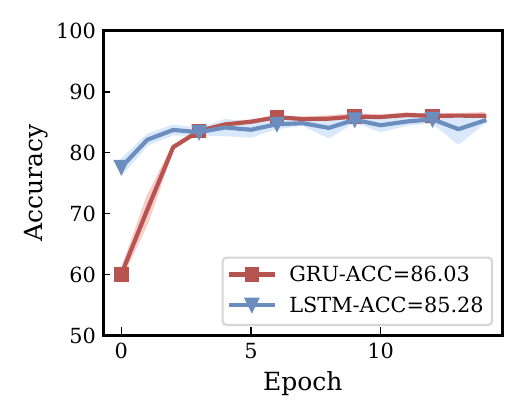}}
\subfigure{\includegraphics[width=45mm, height=40mm]{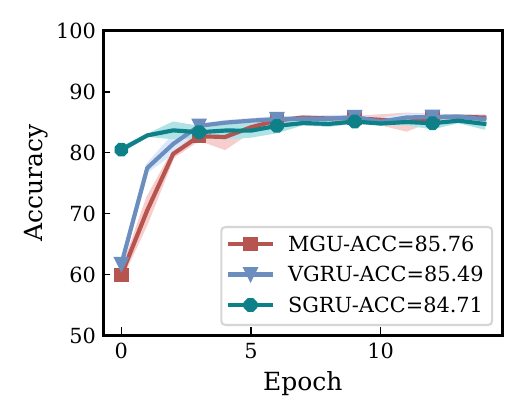}}
\subfigure{\includegraphics[width=45mm, height=40mm]{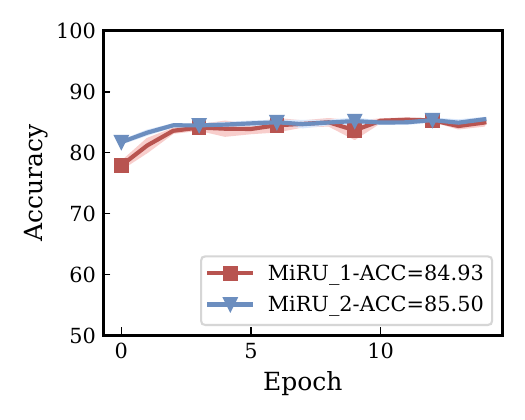}}
\subfigure{\includegraphics[width=45mm, height=40mm]{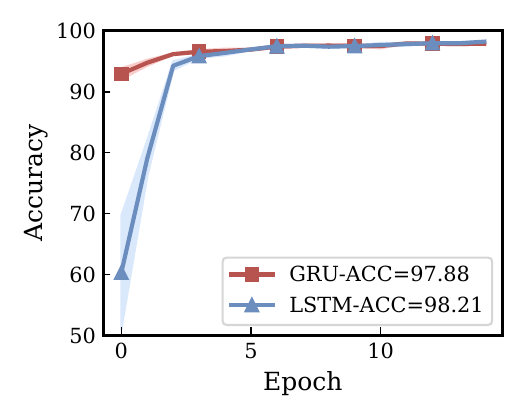}}
\subfigure{\includegraphics[width=45mm, height=40mm]{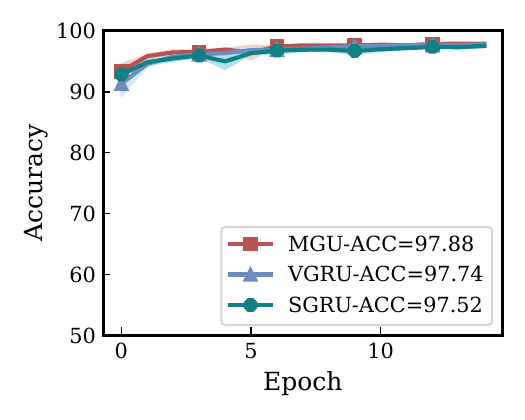}}
\subfigure{\includegraphics[width=45mm, height=40mm]{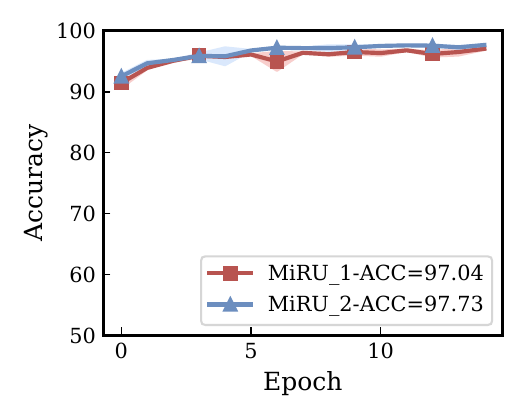}}
\caption{The test accuracy of classifying sequences generated from the IMDB dataset (top row) and MNIST dataset (bottom row) when using the LSTM, GRU, GRU variants (MGU, VGRU, and SGRU), and the proposed MiRU-1 and MiRU-2. {The accuracy is averaged across 5 runs, with the standard deviation represented by shaded light blue and red regions.)}.\vspace{12pt}}
\label{accur_th}
\end{figure*}

\section{Results and Analysis}
\subsection{Classification Accuracy}
\begin{figure*}[h!t]
\centering
\includegraphics[width=135mm, height=40mm]{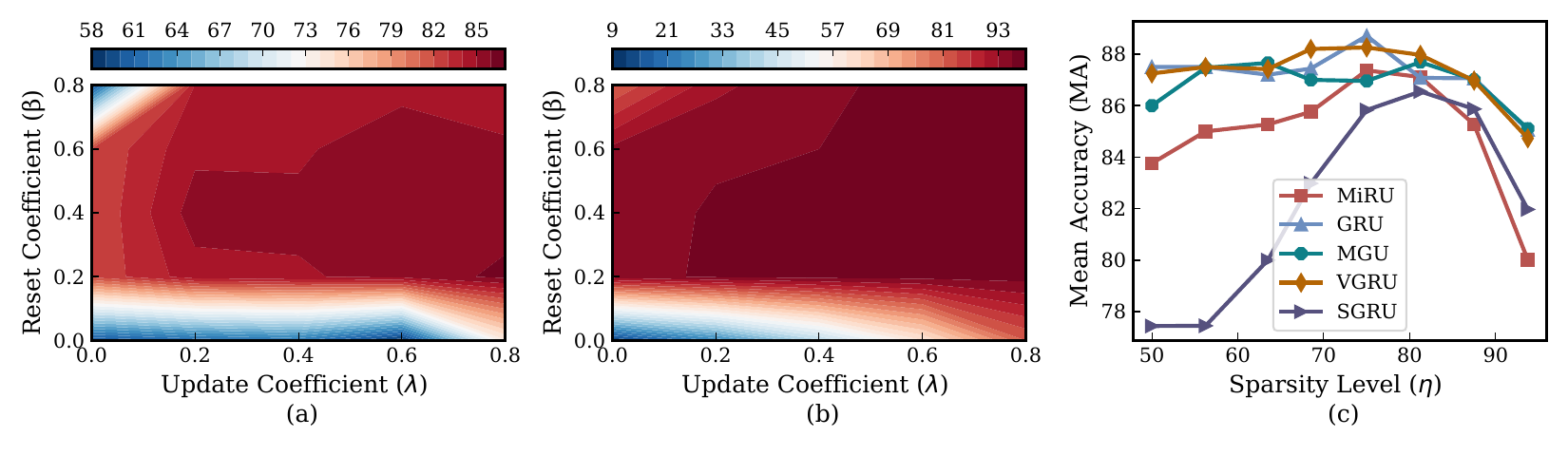}
\caption{{The impact of tuning the reset and update coefficients on test accuracy when classifying sequences generated from (a) IMDB and (b) MNIST datasets. (c) The impact of the sparsity level of the hidden layer neuronal activation on the mean accuracy when learning multiple tasks constructed from permuted MNIST. Best viewed in color.}}
\label{hyper_tuning}
\end{figure*}

The performance of the proposed MiRU models is quantified by classifying sequences generated from the IMDB and MNIST datasets. The results are compared with the LSTM, GRU (baseline architecture), and its variants (MGU, VGRU\footnote{Several VGRU models are introduced in~\cite{dey2017gate}. We have chosen the model that offers best performance.}, and SGRU). Each model is set up to have 128 hidden neurons and output neurons corresponding to the number of classes. The learning rate is set to 0.001 and RMSProp is used as an optimizer to tune the network parameters during learning.~\fig{accur_th} illustrates the mean test accuracy and standard deviation (shaded regions) of all models recorded for 15 epochs for both IMDB (top-row) and MNIST (bottom-row) datasets. It can be seen that although the standard GRU seems to offer comparable results to LSTM and better performance as compared to other models, the reduction in performance is marginal, estimated to be $\sim$0.53\% (IMDB) and $\sim$0.15\% (MNIST) as compared to MiRU-2. This reduction in performance has been noticed in previous variants of GRUs as well. The figure also shows that model convergence is highly impacted by the dataset. For instance, the standard GRU seems to offer faster convergence with marginal differences when classifying MNIST, whereas MiRU-1 and MiRU-2 demonstrate faster convergence when classifying IMDB.{~\fig{hyper_tuning} illustrates the impact of tuning the reset and update coefficients on the test accuracy of the MiRU model for both IMDB (a) and MNIST (b) datasets. It can be observed that eliminating the reset coefficient ($\beta = 0$) causes approximately $12\% - 25\%$ and $17\% - 73\%$ deterioration in MiRU's performance when classifying IMDB and MNIST datasets, respectively. This occurs because eliminating the reset coefficient promotes resetting and adaptation to new inputs while ignoring prior hidden states. Conversely, increasing the reset coefficient to fall between 0.2-0.6 seems to be sufficient to strike a balance between preserving past information and incorporating new data. Likewise, eliminating the update coefficient ($\lambda = 0$) leads to approximately $3\% - 26\%$ and $3\% - 17\%$ degradation when classifying IMDB and MNIST datasets, respectively. This degradation primarily arises from the inability of the MiRU unit to effectively form long-term dependencies. Increasing the update coefficient, on the other hand, enables incorporating new information and retaining past history, thereby forming both short- and long-term dependencies.}

\begin{table*}[h!t]
\setlength\tabcolsep{4pt}
\footnotesize
\caption{{The mean accuracy and standard deviation, averaged across 5 runs, of the proposed MiRU, the standard GRU and its variants after training sequentially on 5 tasks from permuted-MNIST while considering the following scenarios: }}
\label{p_MNIST_1}
\subfigure[Using no continual learning
mechanisms.]{%
~
\label{p_MNIST_1}%
\begin{tabular}{@{}c|ccccc @{}}
\toprule
\textbf{Task}  &  \textbf{GRU} &  \textbf{MiRU-2} & \textbf{MGU}    & \textbf{VGRU} & \textbf{{SGRU}}\\ 
\midrule
Task-1 & 10.42$\pm$0.75 & 13.36$\pm$2.91 & \textbf{16.27$\pm$4.17} & 11.47$\pm$1.45 & {13.05$\pm$1.50}\\
Task-2 & 13.63$\pm$3.59 & 16.08$\pm$3.36 & \textbf{17.73$\pm$2.92} & 14.53$\pm$6.63 & {10.41$\pm$2.39}\\
Task-3 & 19.37$\pm$7.09 & 16.79$\pm$4.36 & \textbf{22.46$\pm$3.82} & 18.56$\pm$8.02 & {15.73$\pm$2.89}\\
Task-4 & 42.55$\pm$9.48 & 40.59$\pm$7.52 & \textbf{44.98$\pm$7.58} & 30.18$\pm$1.06 & {27.70$\pm$1.17}\\
Task-5 & 91.63$\pm$6.10 & \textbf{93.31$\pm$0.47} & 76.02$\pm$0.20 & 83.41$\pm$1.06 & {75.63$\pm$15.78}\\
\midrule
Mean & 35.52$\pm$5.40 & \textbf{36.02$\pm$3.72} & 35.49$\pm$3.74 & 31.63$\pm$7.04 & {28.50$\pm$4.75} \\
\midrule
\midrule
\end{tabular}}%
\hfill%
\subfigure[Using replay with buffer-size set to 3750 per task.]{%
\label{p_MNIST_3}%
\begin{tabular}{@{}c|ccccc @{}}
\toprule
\textbf{Task}  &  \textbf{GRU} &  \textbf{MiRU-2} & \textbf{MGU}    & \textbf{VGRU} & \textbf{{SGRU}}\\ 
\midrule
Task-1 & 71.09$\pm$31.33 & 84.09$\pm$0.91 & {84.12$\pm$0.92} & 71.43$\pm$25.21  & \textbf{{85.67$\pm$1.83}}\\
Task-2 & 75.49$\pm$24.92 & {83.29$\pm$1.06} & 82.61$\pm$1.18 & 79.44$\pm$5.15 & \textbf{{85.08$\pm$2.81}}\\
Task-3 & 75.49$\pm$28.51 & 84.97$\pm$0.77 & \textbf{85.14$\pm$1.41} & 71.29$\pm$29.06 & {85.21$\pm$2.87}\\
Task-4 & 74.38$\pm$31.15 & \textbf{87.04$\pm$0.66} & 86.26$\pm$1.18 & 76.55$\pm$23.52 & {86.15$\pm$2.59}\\
Task-5 & 91.26$\pm$5.34 & \textbf{93.52$\pm$0.73} & 92.16$\pm$0.72 & 87.52$\pm$11.96 & {92.02$\pm$3.41}\\
\midrule
Mean & 77.41$\pm$24.25 & \textbf{86.83$\pm$0.73} & 86.59$\pm$0.94 & 75.68$\pm$23.55 & {86.82$\pm$2.70}\\
\midrule
\midrule
\end{tabular}}%

\subfigure[Using replay with
buffer-size set to 1875 per task.]{%
\label{p_MNIST_2}
\begin{tabular}{@{}c|ccccc @{}}
\toprule
\textbf{Task}  &  \textbf{GRU} &  \textbf{MiRU-2} & \textbf{MGU}    & \textbf{VGRU} & \textbf{{SGRU}}\\ 
\midrule
Task-1 & \textbf{79.58$\pm$2.29} & 78.47$\pm$0.77 & 65.23$\pm$18.19 & 79.42$\pm$2.60 & {79.45$\pm$2.80}\\
Task-2 & 80.53$\pm$4.05 & 77.35$\pm$1.61 & 77.08$\pm$4.44 & {81.02$\pm$2.26} & \textbf{{81.42$\pm$3.49}}\\
Task-3 & 78.17$\pm$15.16 & 79.32$\pm$1.54 & 74.32$\pm$10.96 & \textbf{81.15$\pm$3.39} & {79.03$\pm$6.08}\\
Task-4 & 85.53$\pm$2.16 & 83.18$\pm$1.09 & 82.18$\pm$5.59 & \textbf{85.74$\pm$2.17} & {85.04$\pm$3.89}\\
Task-5 & \textbf{93.86$\pm$0.88} & 93.56$\pm$0.62 & 91.25$\pm$1.77 & 93.30$\pm$0.46 & {88.81$\pm$8.09}\\
\midrule
Mean & 83.53$\pm$5.05 & 82.38$\pm$1.13 & 78.01$\pm$8.19 & \textbf{83.71$\pm$2.24} & {82.75$\pm$4.87}\\
\midrule
\midrule
\end{tabular}}%
\hfill%
\subfigure[Using replay with buffer size
of 1875 and global inhibition.]{%
\label{p_MNIST_4}
\begin{tabular}{@{}c|ccccc @{}}
\toprule
\textbf{Task}  &  \textbf{GRU} &  \textbf{MiRU-2} & \textbf{MGU}    & \textbf{VGRU} & \textbf{{SGRU}}\\ 
\midrule
Task-1 & \textbf{86.08$\pm$0.78} & 83.28$\pm$0.42 & 84.31$\pm$0.67 & 85.44$\pm$0.27 & {81.99$\pm$0.68}\\
Task-2 & \textbf{86.16$\pm$0.72} & 84.94$\pm$0.70 & 85.14$\pm$0.64 & 85.35$\pm$0.49 & {83.28$\pm$0.84}\\
Task-3 & \textbf{87.81$\pm$0.72} & 85.68$\pm$0.47 & 84.45$\pm$4.71 & 87.20$\pm$1.00 & {84.04$\pm$1.10}\\
Task-4 & \textbf{90.38$\pm$0.73} & 88.63$\pm$0.94 & 88.85$\pm$0.31 & 87.74$\pm$3.22 & {86.30$\pm$1.08}\\
Task-5 & 94.17$\pm$0.29 & 93.95$\pm$0.26 & 93.64$\pm$0.14 & \textbf{94.18$\pm$0.21} & {93.35$\pm$0.34}\\
\midrule
Mean & \textbf{88.92$\pm$0.65} & 87.30$\pm$0.56 & 87.28$\pm$1.29 & 87.98$\pm$1.04 & {85.79$\pm$0.81}
 \\
\midrule
\midrule
\end{tabular}}%
\end{table*}

\begin{figure*}[h!t]
\centering
\includegraphics[width=135mm, height=75mm]{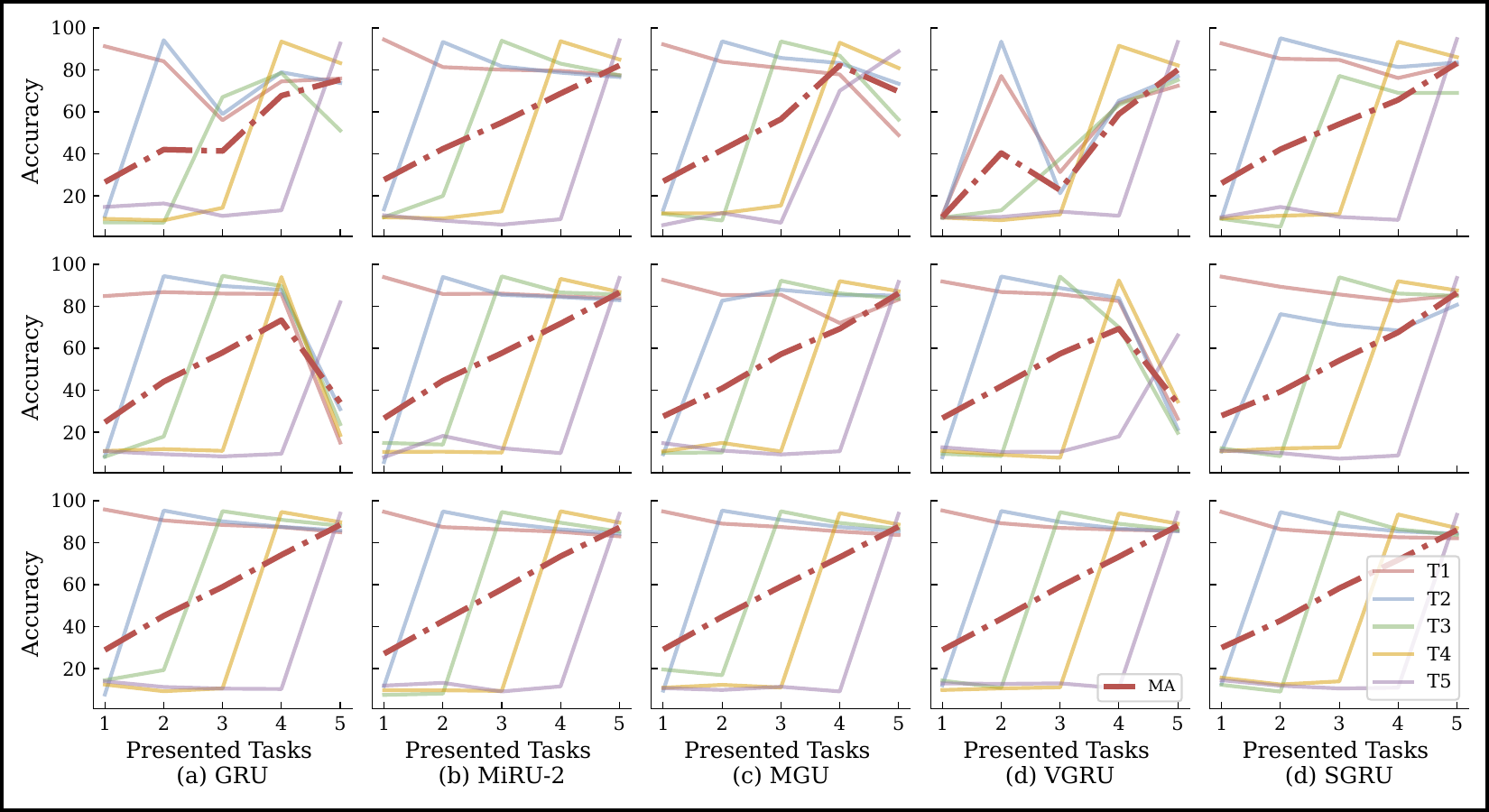}
\caption{The test accuracy (worst-case) and the mean (dotted line) across 5 tasks generated from permuted-MNIST for the MiRU-2, GRU, MGU, VGRU, and SGRU models under the following scenarios: (first-row) replay buffer of size 1875 per task, (second row) replay buffer of size 3750 per task, and (third-row) replay buffer of size 1875 per task with global inhibition. Best viewed in color.}
\label{accur_thx}
\end{figure*}

\subsection{Continual Learning and Catastrophic Forgetting}
In the context of continual learning and catastrophic forgetting, the performance of all models is quantified as they learn a series of tasks formed by sequences generated from five different sets of permuted MNIST. Here, the domain-incremental learning scenario is considered, i.e. the tasks share the same output layer and the task identity is not provided at inference time.~\tb{p_MNIST_1}-(d) report the accuracy averaged across five runs for the individual tasks and the mean accuracy (MA) across the entire set of tasks, given by~\eq{mean_acc}, where $R_{T,i}$ denotes the accuracy of the model on task $i$ after learning $T$ tasks sequentially.
\begin{equation}
    MA = \frac{1}{T}\sum\limits^T_{i=1} R_{T,i}
\label{mean_acc}
\end{equation}

In this work, we consider four possible scenarios to explore continual learning in RNNs. Initially, no continual learning mechanism is applied and each model (baseline) is sequentially trained on all tasks (see~\tb{p_MNIST_1}). The training between tasks is mediated by an inference phase to verify the operation of the network on current and preceding tasks. It can be seen that most models offer high performance on the last task $\mathcal{D}_T$ and catastrophically forget previous tasks \{$\mathcal{D}_1, \mathcal{D}_2, ..., \mathcal{D}_{T-1}$\}. This issue can be mitigated or overcome by rehearsing the network on previous tasks while learning the current task. Rehearsal is typically performed using replay buffers to store samples from the history. {Here, we consider selecting one example (buffer size of 1875 per task) and two examples (buffer size of 3750 per task) from each mini-batch during training to preserve the learned knowledge. The selection is carried out using reservoir sampling with no fix limits on the number of stored examples per class.}

When using 1875 samples per task, which is the second scenario, the RNN models no longer experience catastrophic forgetting and are able to handle previously learned tasks with a reasonable drop in performance, see~\tb{p_MNIST_2}. The table also illustrates that despite most models seem to offer comparable mean accuracy, high fluctuations in performance are observed in the standard GRU and its variants after learning multiple tasks unlike MiRU-2.~\fig{accur_thx}-(first-row) shows the worst case among 5 runs for all the tested models. It is evident that the standard GRU and its variants occasionally struggle to learn multiple tasks, unlike MiRU-2, which shows a robust learning behavior with its mean accuracy consistently improving as the model learns new tasks. In the third scenario, in which we extend the replay buffer size to 3750, we notice a substantial increase in the fluctuation, especially in GRU and VGRU (see~\tb{p_MNIST_3} and \fig{accur_thx}-(second-row)). Besides the increase in fluctuation, we witness an improvement in performance in some models, but to varying degrees, with MiRU-2 outperforms all the models. The improvement in performance when extending the buffer size is attributed to three factors: i) increase the number of samples drawn from past events, leading to more balanced parameter tuning, i.e. the network training will not bias towards recent experiences, ii) ameliorate the network generalizability via broadening the number of experiences stored from the past, iii) make the training more stable via interleaving old and current experiences. While increasing the buffer size of the replay seems to boost network performance, it is not a feasible option in numerous cases especially when dealing with edge devices with stringent constraints. Therefore, in the forth scenario, we suggest using the replay with the global inhibition inspired by biology to reduce the buffer size. {To determine the optimal sparsity level for global inhibition, we test various sparsity levels ranging from 50\% to 94\% and observe their impact on the mean accuracy (MA) across all models. The results, as illustrated in~\fig{hyper_tuning}-(c), indicate that a sparsity level of 75\% yields the best performance across most models. Thus, we select a sparsity level of 75\% for our work.} 

\begin{figure}[!t]
    \centering
    \includegraphics[width=135mm, height=40mm]{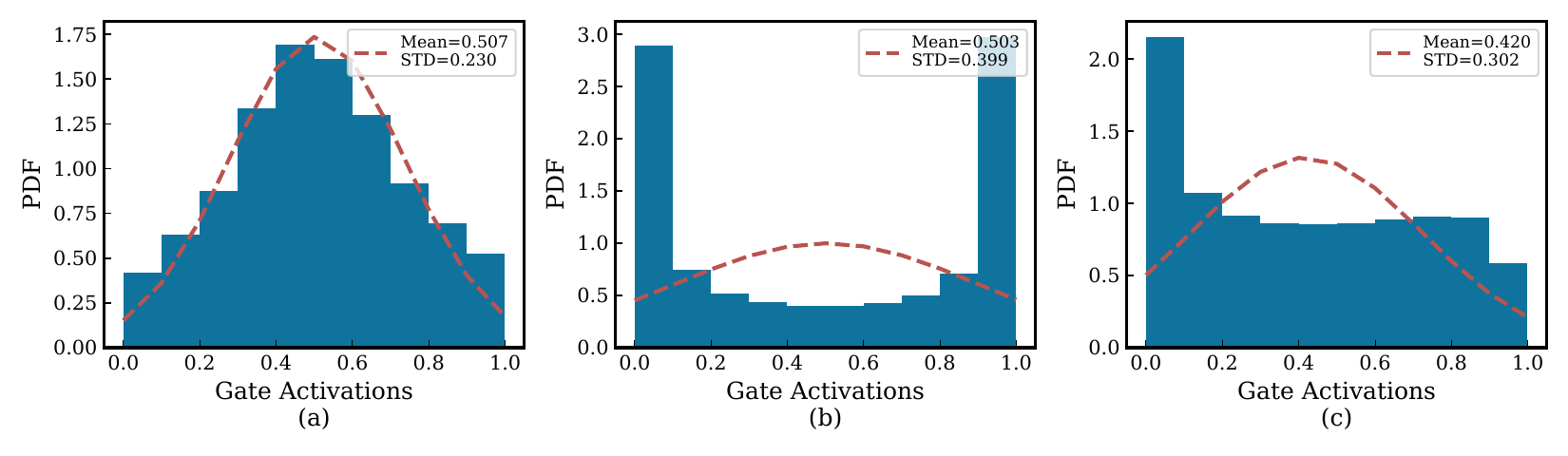}
    \caption{The distribution of the GRU forget gate activation: (a) at initialization, (b) after learning the first task without global inhibition, and (c) after learning the first task with global inhibition.}
    \label{gate_activties}
\end{figure}

It important to mention her that using global inhibition stimulates the competition among neurons in the hidden layer, allowing only neurons with the highest activations to represent the input. This eventually reduces the overlap between tasks and minimizes the risk of overwriting information learned in previous tasks. Furthermore, it prevents overfitting and enhances network generalization to new tasks as the network is forced to learning more abstract representation using a subset of neurons rather than relying on all neurons to learn a given task. At the gate level, global inhibition seems to play a significant role in controlling the distribution of gate activities during multitask sequential learning. For instance,~\fig{gate_activties} demonstrates the distribution of the GRU update gate before and after learning, both in the absence and presence of global inhibition. Initially, it can be observed that the gate follows a normal distribution. After learning first task, the distribution shifts to U-shape, indicating most of the activations of the update gates in the hidden layer are either 0 or 1, see~\fig{gate_activties}-(b). This behavior makes the gates prone to saturation property, hindering their gradient-based learning in the future~\cite{gu2020improving}. Such a problem is significantly mitigated when using global inhibition, as the update gate activities become more evenly distributed between 0 and 1 and less concentrated at the boundaries\footnote{The distribution of forget gate activations, both with and without global inhibition, is maintained even after learning subsequent tasks.} as shown in~\fig{gate_activties}-(c).~\tb{p_MNIST_4} and \fig{accur_thx}-(third-row) illustrate the significant improvement in all models performance when using both replay and global inhibition. Besides the improvement in performance, stable performance across tasks is also achieved.

\begin{figure}[h!t]
    \centering
    \includegraphics[width=70mm, height=45mm]{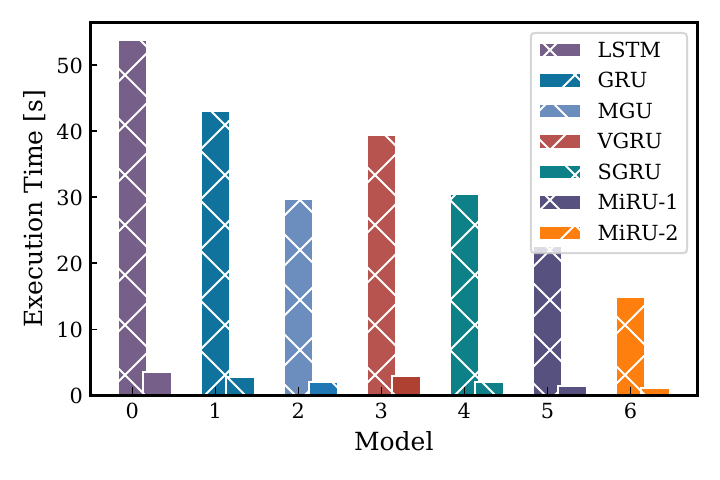}
    \caption{The training time (left-sided bars) and inference time (right-sided bars) of the LSTM, GRU, MGU, VGRU, SGRU, and MiRU when processing sequences generated by MNIST training and test sets.}
    \label{execution_time}
\end{figure}

\subsection{Execution Time}
RNNs adaptation speed and energy efficiency are highly controlled by the execution time, training, and inference. In this work, the averaged training and inference times of the proposed MiRU, LSTM, GRU, MGU, VGRU, and SGRU are estimated over 10 epochs using sequences generated from MNIST images. The training time is measured for the entire training set of 60,000 images, whereas the test set of 10,000 images is used to estimate the inference time. Both sets are processed in mini-batches of size 32. All models are implemented in Python using Pytorch and run on a personal MacBook Pro laptop. The laptop is equipped with an i7 2.6GHz 6-core processor, 16GB 2400MHz DDR4 RAM, and 1TB SSD.~\fig{execution_time} shows the training time (left-sided bars) and inference time (right-sided bars) for each model. It can be seen that our model, MiRU-2, training and inference times are estimated to be 14.83s and 1.13s, respectively. Compared to other models, it demonstrates significant improvements. For example, compared to the standard GRU, it achieves 2.90$\times$ and 2.49$\times$ faster training and inference times.

\subsection{Resource Utilization}
Enabling time-series data processing on edge devices is contingent on available resources, the size of the RNN model to be ported on the edge device, and the complexity of the trained model~\cite{zyarah2024esn}. Edge devices with limited resources demand compact RNN models that can be easily adapted and trained, especially when used in continual learning applications. Given the complexity of LSTM and GRU, running either one on edge devices seems to be challenging while enabling in situ learning. Therefore, we introduce MiRU-2 that offers approximately a three-fold decrease in the number of parameters and activation functions compared to GRU; see~\tb{resources}. When it comes to computational cost, MiRU-2 seems to be computationally light compared to GRU and its variants. Quantitatively, the computational cost is estimated during the forward and backward passes while processing sequences generated from MNIST dataset. However, due to differences in the dominant operations, the forward pass is evaluated in terms of MAC operations, whereas the backward pass\footnote{The backward pass is dominated by outer products, element-wise multiplications, and additions.} is evaluated in terms of multiplications and additions.~\tb{resources} 
reports $\sim$4.2$\times$, $\sim$2.6$\times$, $\sim$3.88$\times$, and $\sim$1.011$\times$ 
reduction in the number of multiplications compared to the standard GRU, MGU, VGRU, and SGRU. For additions, a similar trend of reduction is observed. and estimated to be $\sim$24.72$\times$, $\sim$12.81$\times$, $\sim$24.72$\times$, and $\sim$1.636$\times$, respectively. When it comes to MAC operations, they are reduced by $\sim$2.81$\times$, $\sim$1.934$\times$, $\sim$2.371$\times$, and $\sim$1.011$\times$, respectively.

\begin{table*}[ht!]
\footnotesize
\setlength\tabcolsep{2 pt}
\caption{The number of parameters, activation functions, MAC operations required to perform a single forward and backward pass of MNIST sequences, and the inference energy consumption for the the proposed MiRU, standard GRU RNN, and other variants presented in the literature. For all models, the network size is set to 28$\times$128$\times$10 and trained using the same optimizer and learning rate.}
\begin{center}
\begin{tabular}{lcccccc}
\toprule   
{\textbf{Model}}   &  {\textbf{No. of Parameters}}  & {\textbf{Forward (MAC)}} & {\textbf{Act. Functions}} & {\textbf{Backward (Mul)}} & {\textbf{Backward (Add)}}& {\textbf{ Energy (\boldmath{$\mu$}J)}}
\\ 
\midrule
{GRU}  & 61,696   & 60,160   & 394 & 96,768      & 34,816 &  0.391 \\
{MiRU-1}  & 41,472   & 41,477   & 266 &59,776      & 17,920  &  0.263\\ 
{MiRU-2}  & 21,386   & 21,376   & 138 & 23,040      & 1,408  &  0.136\\ 
MGU  & 41,472   & 41,344   & 266 & 59,904  & 18,048   &  0.261   \\ 
VGRU & 54,016   & 50,688   & 394 & 89,600      & 34,816  &  0.342  \\ 
SGRU  & 21,898   & 21,632   & 394 & 23,305      & 2,304  &  0.139  \\ 
{LSTM}  & 81,674   & 81,152   & 650 & 121,216 & 38,096  &  0.519  \\ 
\midrule
\bottomrule
\end{tabular}
\end{center}
\label{resources}
\end{table*}


\begin{table}[h!t]
\footnotesize
\centering
\caption{{{The energy consumption of the core computational units, activation functions, and storage units.} }}
{%
\begin{tabular}{@{}lcc @{}}
\toprule
\textbf{Unit}  &  \textbf{Bit Precision} &  \textbf{{Energy (pJ)}}\\ 
\midrule
Multiplier & 16-bit & 0.11 \\
Adder & 16-bit & 0.013 \\
Accumulator & 24-bit & 0.235 \\
MAC & 24-bit & 0.376\\
Tanh Activation & 16-bit & 0.326 \\
Sigmoid Activation & 16-bit & 0.228\\
\midrule
SRAM read & 16-bit & 5.893  \\
SRAM write & 16-bit & 6.628  \\
\midrule
\midrule
\end{tabular}}%
\label{energy}
\end{table}

\subsection{Energy Consumption}
The inference energy consumption of the proposed MiRU, standard GRU, and its variants is analytically estimated based on the workload of the computational units and the access of the storage when predicting sequences. All core computational units are designed in Verilog HDL and synthesized using the Synopsys design compiler (DC) under 65nm technology node. The power consumption is then estimated using Synopsys PrimeTime-PX tools with the system clock set to 100MHz. When it comes to storage, SRAM is considered and its energy consumption during read and write operations is estimated using the HP Cacti V7.0 tool.

It is worth mentioning that all computational units, along with activation functions, are modeled in Verilog using mixed fixed-point precision based on the design proposed in~\cite{zyarah2023reservoir}. This is to ensure reasonable performance with a negligible gap between the software and hardware models.~\tb{energy} illustrates the energy consumption of computational, activation, and storage units. One may observe that the activation units, approximated using pice-wise functions, consume energy marginally different from the computational units, unlike the storage unit. The energy consumed by the SRAM when retrieving or updating network parameters is significantly higher. Thus, reducing the number of parameters plays an important role in reducing the overall energy consumption of the proposed MiRU RNN compared to its counterparts, see~\tb{resources}.

\section{Conclusions}
In this work, the minion recurrent unit (MiRU) for recurrent neural networks is introduced. The proposed model utilizes scaling coefficients rather than conventional gates to control the hidden state update dynamics and the history. This results in a compact model that uses 2.88$\times$ fewer parameters than the standard GRU and is more cost-effective in terms of computation. Specifically, it requires $\sim2.81\times$ fewer MAC operations during the forward pass and $\sim4.2\times$ and $\sim24.72\times$ fewer multiplications and additions during the backward pass. The reduction in parameters and computations is also reflected in training time and inference time. MiRU achieves 2.90$\times$ and 2.49$\times$ faster training time and inference time compared to standard GRU. {It also demonstrates $\sim2.87\times$ better energy efficiency than the standard GRU}. We evaluated the effectiveness of the proposed MiRU for sequence classification generated from IMDB and MNIST datasets, and it is found that MiRU offers an accuracy comparable to GRU and its variants. Furthermore, we explore the ability of MiRU, GRU, MGU, VGRU, and SGRU to perform domain-incremental learning enabled by replay and global inhibition. When tested with permuted MNIST, MiRU offers stable performance when using only replay, unlike its counterparts, which suffer from fluctuation in performance. However, this issue is overcome when using replay and global inhibition. In general, MiRU demonstrates compactness, computational efficiency, and performance stability, making it a promising candidate for deployment on resource-constrained edge devices, enabling continuous learning and sequence classification.


\bibliographystyle{ACM-Reference-Format}
\bibliography{References}
\end{document}